\documentclass[runningheads]{llncs}

\usepackage[mobile]{eccv}

\usepackage{eccvabbrv}

\usepackage{graphicx}
\usepackage{booktabs}
\usepackage{mathrsfs}
\usepackage{multirow}
\usepackage{makecell}
\usepackage{colortbl}
\usepackage{booktabs}
\usepackage{array}
\usepackage{newfloat}
\usepackage{listings}
\usepackage{makecell}
\usepackage[symbol]{footmisc}
\usepackage[accsupp]{axessibility}  
\newcommand{\STAB}[1]{\begin{tabular}{@{}c@{}}#1\end{tabular}}
\usepackage{algorithm}
\usepackage{algpseudocode}

\usepackage[pagebackref,breaklinks,colorlinks,citecolor=eccvblue]{hyperref}

\usepackage{orcidlink}

\begin{document}

\title{Token Compensator: Altering Inference Cost of Vision Transformer without Re-Tuning}

\titlerunning{\texttt{ToCom}: Altering Inference Cost of Vision Transformer without Re-Tuning}

\author{Shibo Jie$^{1}$ \and
Yehui Tang$^{2}$ \and
Jianyuan Guo$^{2}$ \and
Zhi-Hong Deng$^{1*}$ \and
Kai Han$^{2*}$ \and
Yunhe Wang$^{2*}$ }

\authorrunning{Shibo Jie \emph{et al}.}

\institute{$^{1}$ State Key Laboratory of General Artificial Intelligence, School of Intelligence Science and Technology, Peking University \\ $^{2}$ Huawei Noah’s Ark Lab}
\maketitle

\begin{abstract}
\footnotetext[1]{Corresponding Author.\quad$^\dag$\url{https://github.com/JieShibo/ToCom}}
Token compression expedites the training and inference of \emph{Vision Transformers} (ViTs) by reducing the number of the redundant tokens, \eg, pruning inattentive tokens or merging similar tokens. However, when applied to downstream tasks, these approaches suffer from significant performance drop when the compression degrees are mismatched between training and inference stages, which limits the application of token compression on off-the-shelf trained models.
In this paper, we propose a model arithmetic framework to decouple the compression degrees between the two stages. In advance, we additionally perform a fast parameter-efficient self-distillation stage on the pre-trained models to obtain a small plugin, called \textbf{To}ken \textbf{Com}pensator (\textbf{\texttt{ToCom}}), which describes the gap between models across different compression degrees. During inference, \texttt{ToCom} can be directly inserted into any downstream off-the-shelf models with any mismatched training and inference compression degrees to acquire universal performance improvements without further training. Experiments on over 20 downstream tasks demonstrate the effectiveness of our framework. On CIFAR100, fine-grained visual classification, and VTAB-1k, \texttt{ToCom} can yield up to a maximum improvement of 2.3\%, 1.5\%, and 2.0\% in the average performance of DeiT-B, respectively.$^\dag$
\end{abstract}

\section{Introduction}
\emph{Vision Transformers} (ViTs)~\cite{vit} have achieved remarkable success in various fields of computer vision, including image classification~\cite{deit}, object detection~\cite{vitdet,vit-adapter}, semantic segmentation~\cite{segmenter}, \etc. However, with the rapid growth in the scale of ViTs, the increasing computational cost has become a pressing issue. Consequently, a large number of efforts are focusing on accelerating the training and inference of ViTs~\cite{compression1,compression2,compression3,compression4,compression5,compression6}. The characteristic of ViTs lies in their capacity to accommodate a variable number of input tokens. Thus, beyond the conventional techniques widely utilized in convolutional neural networks such as model pruning, quantization, and distillation, recent researches suggest the acceleration of ViTs through token compression, such as pruning inattentive tokens~\cite{evo-vit,evit,a-vit,ats} or merging similar tokens~\cite{tome,token-pooling,tps}.

Token compression techniques offer distinct advantages. In comparison to techniques like pruning and distillation, some token compression approaches 
(\eg, ToMe~\cite{tome}) can be applied in a zero-shot manner to off-the-shelf models or utilized for accelerating training; unlike quantization, token compression methods do not necessitate support for low-precision operators. Furthermore, token compression methods operate orthogonally to the aforementioned other techniques, rendering them widely applicable in ViTs.

\begin{figure}[tb]
  \centering
  \includegraphics[height=4.0cm]{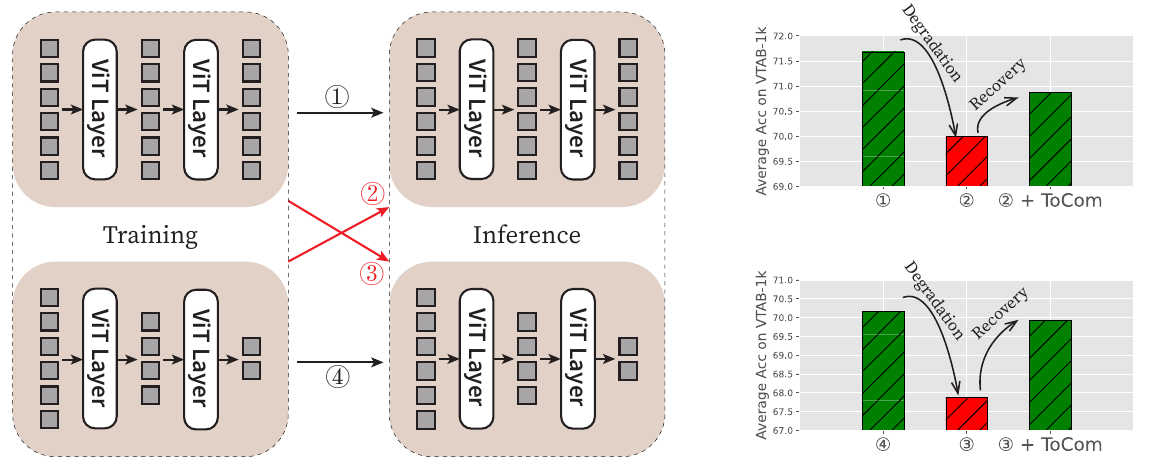}
  \caption{\textbf{\emph{Left:}} Previous token compression methods focus on scenario when training and inference compression degrees are consistent (\textcircled{1} and \textcircled{4}), but not adequately address the performance of models when these degrees differ (\textcircled{2} and \textcircled{3}). \textbf{\emph{Right:} }Performance of token compression significantly degrades when compression degrees in training and inference are not equal. After applying our \texttt{ToCom} without training, the performance is recovered.}
  \label{fig:intro}
\end{figure}

However, we observe the following drawbacks in token compression when applied to downstream tasks. {Firstly}, although some token compression techniques can be applied to off-the-shelf models, they often lead to significant performance degradation. {Secondly}, even if token compression is only applied during training to expedite the process, and tokens are not compressed during inference, the model's performance still falls below models trained without token compression. In summary, \emph{when there is inconsistency in the token compression degrees between training and inference stages, the performance of models is suboptimal}, as illustrated in Fig.~\ref{fig:intro}. This limitation restricts  its application across various scenarios. For example, if we aim to dynamically adjust the computation costs of deployed models based on server load, we would need to train a model for each  candidate of computation costs and each downstream task to achieve optimal performance, resulting in significant training and storage overheads. Additionally, if we intend to accelerate training with minimal loss in model inference performance, token compression in training-time only may lead to obvious performance drop.

In this paper, we point out that models fine-tuned under different token compression degrees exhibit a certain gap at the parameter level, causing performance degradation when altering compression degrees during inference. Also, we observe that this gap can be transferred across different downstream datasets. Motivated by this, we propose \textbf{To}ken \textbf{Com}pensator (\texttt{ToCom}), a pre-trained plugin designed to decouple the token compression degrees between training and inference, to address the aforementioned challenges. \texttt{ToCom} is a parameter-efficient module that only contains a negligible number of parameters, which describes the gap between models with different compression degrees. To obtain \texttt{ToCom}, we train it on pre-training datasets with a fast self-distillation process among different compression degrees. To elaborate, both the teacher model and the student model are the same frozen pre-trained model, with the student model incorporating \texttt{ToCom}. Different compression degrees are randomly assigned to the teacher and student models in each step, while \texttt{ToCom} learns the gap between them through distillation. Moreover, we allocate different subsets of \texttt{ToCom} parameters for different compression degree pairs, enabling \texttt{ToCom} to adapt to various compression degree pairs through a single training process.

During inference, we directly integrate \texttt{ToCom} into off-the-shelf models fine-tuned on downstream tasks without any further training. By selecting the subsets of \texttt{ToCom} parameters, the fine-tuned model can be directly applied to various token compression degrees and achieve on-par performance comparable to the case when training and inference compression degrees are consistent. Importantly, \texttt{ToCom} only needs to be pre-trained once and can be applied to models fine-tuned on arbitrary downstream datasets with arbitrary token compression degrees, thereby enable any single off-the-shelf model to handle dynamic latency constraints without parameter modification. 

We conduct experiments on over 20 datasets across various settings of compression degree. Experimental results demonstrate that \texttt{ToCom}, as a plug-and-play module, can effectively decouple the compression degree between training and inference for token compression. For example, on VTAB-1k benchmark, \texttt{ToCom} can yield up to a maximum improvement over ToMe of 2.0\% in the average performance of DeiT-B, as shown in Fig.~\ref{fig:intro}. \texttt{ToCom} can also be applied to models with different scales or models pre-trained with different  objects, or utilized to enhance various token compression methods including token merging and token pruning.

\section{Related Work}
\subsubsection{Token Compression}
Token compression aims to eliminate redundancy by reducing the number of tokens, thereby expediting ViTs. It primarily encompasses two directions: token pruning and token merging. Token pruning involves assessing token importance through defined metrics and removing unimportant tokens accordingly. PS-ViT~\cite{ps-vit} proposes a top-down paradigm to estimate the importance of each token. DynamicViT~\cite{dynamicvit} fine-tunes lightweight subnetwork to evaluate tokens. EViT~\cite{evit} and Evo-ViT~\cite{evo-vit} use the attention score of \texttt{[CLS]} tokens as a metric of attentiveness, and optionally pools all the pruned tokens into one. A-ViT~\cite{a-vit} and ATS~\cite{ats} also adjust the pruning rate based on the complexity of the input image, but their dynamic token length is not well-supported by current computation framework. SuperViT~\cite{supervit} trains a single ViT to support various token keeping rates.

On the other hand, token merging reduces token quantity by merging similar tokens. Token Pooling~\cite{token-pooling} uses $k$-means for token clustering. ToMe~\cite{tome} proposes a Bipartite Soft Matching (BSM) algorithm to gradually merge similar tokens. TPS~\cite{tps} achieves token merging by performing pruning first, and fusing the reserved token and pruned tokens. CrossGET~\cite{crossget} propose Complete-Graph Soft Matching and Cross-Guided Matching for token merging. There are also methods that combine pruning and merging together, such as DiffRate~\cite{diffrate} and PPT~\cite{ppt}.

Among the aforementioned methods, some methods are parameter-free, such as ToMe, ATS, CrossGET, and EViT, and thus can be directly used on off-the-shelf models in inference stage. Moreover, since these methods also reduce the tokens during training, they can be leveraged to accelerate the training stage. However, current research has not thoroughly explored the potential of token compression methods in these aspects.

\subsubsection{Parameter-Efficient Tuning}
\emph{Parameter-Efficient Tuning} (PET) aims to fine-tune pre-trained vision backbones for downstream tasks by adjusting only a small subset of parameters with backbone frozen, including prompt-based methods which append trainable tokens to the sequential inputs of transformers as prompts~\cite{vpt,vpt2,e2vpt}; adapter-based methods which incorporate small adapters into the pre-trained model~\cite{adapter,adaptformer,lora,revisit,consolidator,fact,memvp}; tuning bias parameters only~\cite{bitfit}; altering intermediate features using affine transformation~\cite{ssf}; and matching feature changes in fine-tuning with a small side-network~\cite{side-tuning}.

Previous PET methods focus on adaptation, in other words, using small PET modules to characterize the gap between the pre-trained and fine-tuned models.  In contrast, we take a different perspective in this paper, using the PET modules to describe the gap between models with different compression degrees, serving as a universal compensator.

\subsubsection{Model Arithmetic}

Model arithmetic involves merging, enhancing, or removing specific capabilities between different models through model-level addition or subtraction. \cite{arith1,arith2} show that arithmetic of models can achieve distribution generalization, multi-task learning, unlearning, and domain transfer. Some works on diffusion model  have also utilized model arithmetic to reduce iterative steps~\cite{lcm-lora} or combine the style and object for generating~\cite{ziplora}. However, the application of model arithmetic on token compression has not been explored yet.

\section{Delve into Token Compression}

\subsection{Impact of Compression Degrees}
First, we formulate the token compression method on ViTs. A single layer of ViTs consists of two blocks, namely Multi-Head Self-Attention (MHSA) and Multi-Layer Perceptron (MLP). The layer can be formalized as 
\begin{equation}
    \widetilde{\mathbf{X}}^l=\mathbf{X}^l+\textrm{MHSA}(\textrm{LN}(\mathbf{X}^l)), \quad \mathbf{X}^{l+1}=\widetilde{\mathbf{X}}^l+\textrm{MLP}(\textrm{LN}(\widetilde{\mathbf{X}}^l)),
\end{equation}
in which $\mathbf{X}^l\in\mathbb{R}^{N\times d}$ is the input of the $l$-th layer with length $N$ and dimension $d$, and LN denotes layer normalization.

In this paper, we mainly focus on a representative and state-of-the-art training-free token compression methods, ToMe~\cite{tome}, and then generalize to other methods. ToMe  operates  between MHSA and MLP blocks. It leverages the keys of patch tokens to evaluate their similarity, and merges $r$ similar ones with Bipartite Soft Matching.

\begin{figure}[tb]
  \centering
  \includegraphics[height=3.1cm]{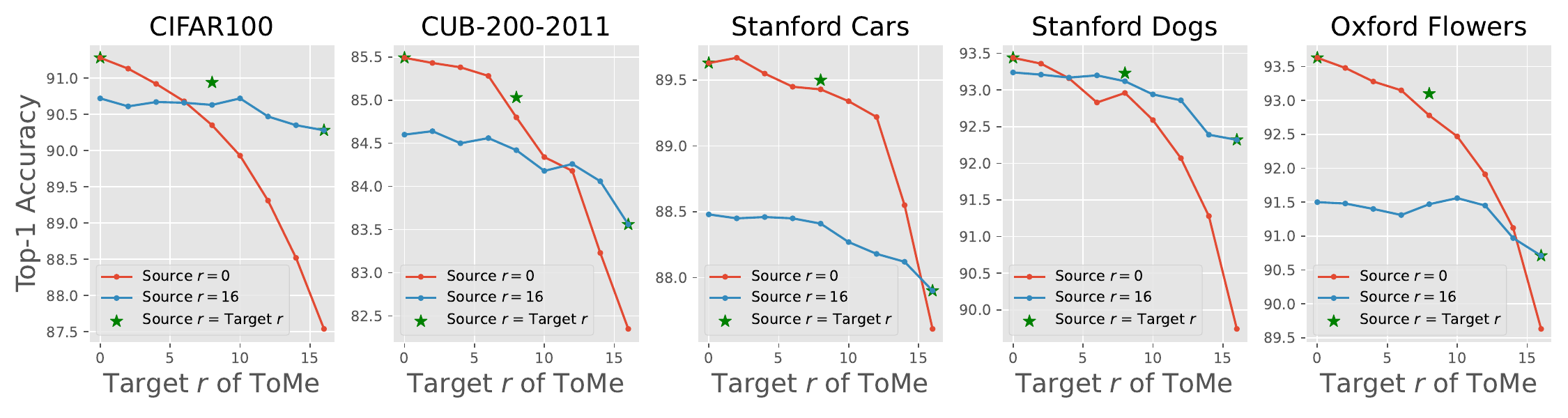}
  \caption{{Performance of ToMe on CIFAR100 and FGVC datasets. }We use DeiT-B as pre-trained backbone. We report performance when source $r\in$ \{0, 16, target $r$\}}.
  \label{fig:rate}
\end{figure}

\begin{figure}[t]
\begin{minipage}[h]{0.49\textwidth}
\centering
\captionof{table}{{Results} of model gaps transfer. We use CIFAR100 as $\mathcal{D_A}$ and FGVC tasks as $\mathcal{D_B}$. The results are evaluated with $r=16$.}
\scalebox{0.8}{
\begin{tabular}{p{2.5cm}<{}p{1.2cm}<{\centering}p{3.2cm}<{\centering}}
\toprule
Dataset $\mathcal{D_B}$&$\mathcal{M}_{0}^{\mathcal{D_B}}$ &$\mathcal{M}_{16}^{\mathcal{D_A}} - \mathcal{M}_0^{\mathcal{D_A}} + \mathcal{M}_0^{\mathcal{D_B}}$ \\\hline
\specialrule{0em}{1pt}{1pt}
CUB-200-2011&82.4&\bf 83.2\\
Stanford Cars&87.6&\bf 87.9\\
Stanford Dogs&89.5&\bf 90.3\\
Oxford Flowers&89.6&\bf 91.2\\
\bottomrule
\label{tab:transfer}
\end{tabular}
}
\end{minipage}
\hfill
  \begin{minipage}[h]{0.49\textwidth}
\centering
\includegraphics[width=0.9\textwidth]{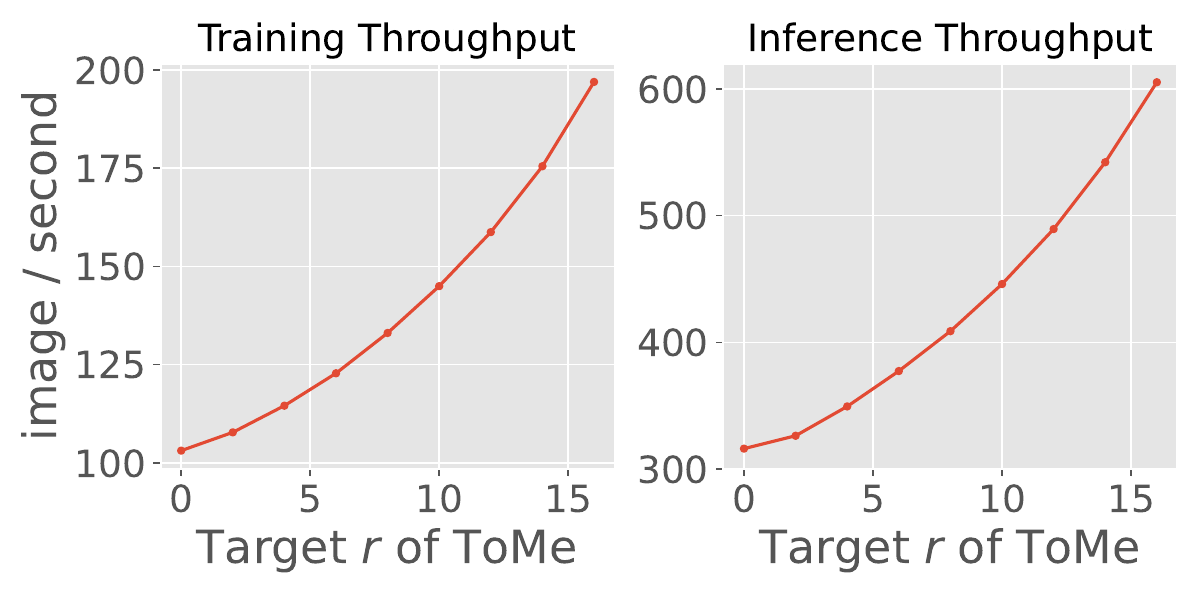}
    \captionof{figure}{{Training and inference throughput of DeiT-B with different $r$ of ToMe. Batch size is 128 and 256 for training and inference, respectively.}} 
    \label{fig:speed}
  \end{minipage}
\end{figure}
In ToMe, the number of merged tokens per layer are considered as hyper-parameters to adjust the throughput of the ViTs, which is usually determined based on inference requirements before training. The more tokens it merges, the faster the models in training and inference, as shown in Fig.~\ref{fig:speed}. However, in real-world scenarios, the compression degrees during training (called \textbf{source degrees}) and inference (called \textbf{target degrees}) may not necessarily be equal. That is, an off-the-shelf model trained at one compression degree might be applied at different compression degrees without retraining. This scenario holds practical significance, such as when utilizing downloaded checkpoints without access to training data or resource for retraining, or dynamically adjusting the compression degree during inference based on server loads. Additionally, in cases where existing computational resources are limited, one may have to use a high compression degree to reduce memory and time overheads during training but restore to a lower compression degree during inference to ensure performance. 

To investigate the performance of token compression methods when there is inconsistency between source and target degrees, experiments are conducted on five downstream datasets~\cite{cifar,cub,cars,dogs,flowers}. As illustrated in the Fig.~\ref{fig:rate}, we  fine-tune DeiT-B~\cite{deit} with ToMe of \(r=0\) and \(16\), and report the performance during inference with \(r=0,2,4,...,16\). We observe that for a specific target degree, the model performs better when the source degree matches it. The greater the disparity between the source and target degrees, the more the performance degradation becomes. 

However, since models trained at lower compression degrees have seen more tokens during training, implying they have encountered a broader range of information than models trained at higher compression degrees, the former ideally should outperform the latter on various target degrees. \emph{This suggests the presence of a gap between models trained at different source degrees, making the transfer between different compression degrees less effective. }

\subsection{Transfer across Tasks}
For the gap between models with different source degrees, we pose a question: \emph{Is this gap transferable across tasks?} More concretely, let $\mathcal{M}_m^{\mathcal{D_A}}$ and $\mathcal{M}_{n}^{\mathcal{D_A}}$ denote models trained with compression degree \( m \) and \( n \) on dataset $\mathcal{D_A}$, and $\mathcal{M}_m^{\mathcal{D_B}}$ and $\mathcal{M}_{n}^{\mathcal{D_B}}$ denote models trained on dataset $\mathcal{D_B}$. If the gap is transferable, we should have
\begin{equation}
    \mathcal{M}_{m}^{\mathcal{D_A}} - \mathcal{M}_n^{\mathcal{D_A}} \approx \mathcal{M}_{m}^{\mathcal{D_B}} - \mathcal{M}_n^{\mathcal{D_B}},
\label{eq:transfer}
\end{equation}
in which $+$ and $-$ are parameter-wise addition and subtraction, respectively. To validate this, we rewrite Eq.~\ref{eq:transfer} as
\begin{equation}
    \mathcal{M}_{m}^{\mathcal{D_A}} - \mathcal{M}_n^{\mathcal{D_A}} + \mathcal{M}_n^{\mathcal{D_B}} \approx \mathcal{M}_{m}^{\mathcal{D_B}},
\label{eq:transfer2}
\end{equation}
which means $(\mathcal{M}_{m}^{\mathcal{D_A}} - \mathcal{M}_n^{\mathcal{D_A}} + \mathcal{M}_n^{\mathcal{D_B}})$ should perform better than $\mathcal{M}_n^{\mathcal{D_B}}$ when evaluated on $\mathcal{D_B}$ with target degree $m$. In other words, the gap $(\mathcal{M}_{m}^{\mathcal{D_A}} - \mathcal{M}_n^{\mathcal{D_A}})$ on $\mathcal{D_A}$ can be transferred to $\mathcal{D_B}$ and help to bridge the gap between $\mathcal{M}_{m}^{\mathcal{D_B}}$ and $\mathcal{M}_n^{\mathcal{D_B}}$. We conduct preliminary experiments on ToMe by using CIFAR100~\cite{cifar} as $\mathcal{D_A}$, $m=16$, and $n=0$. In Table~\ref{tab:transfer}, we show the results when using different FGVC datasets~\cite{cub,cars,dogs,flowers} as $\mathcal{D_B}$. We note that by adding $(\mathcal{M}_{16}^{\mathcal{D_A}} - \mathcal{M}_0^{\mathcal{D_A}})$, the performance of $\mathcal{M}_0^{\mathcal{D_B}}$ on $r=16$ is obviously improved, verifying that \emph{there exist positive transfer of the model gap between two source degrees across different tasks, }which suggests that it is possible to use a universal plugin to model such gap and improve the performance of token compression with unequal source and target degrees on various downstream tasks.

\section{Token Compensator}

\subsection{Arithmetic of Parameter-Efficient Modules}
When the compression degrees during training and inference are unequal, the performance degradation of the model occurs due to the differente behaviors between models with different compression degrees, resulting in distinct local minima in their parameter spaces. To enhance the performance of token compression in such circumstance, especially on the off-the-shelf models, we intend to find a universal plugin to compensate for the gap between models with different compression degrees, assuming that models at two specific compression degrees exhibit similar gaps across different datasets. Supposing we have the plugin $\mathcal{P}_{m\rightarrow n}$ which compensates for the gap between model $\mathcal{M}_m$ {trained with ToMe} $r=m$ and model $\mathcal{M}_{n}$ trained with ToMe $r=n$, if $\mathcal{M}_m$ is off-the-shelf, we can expect that
\begin{equation}
\mathcal{M}_m \oplus \mathcal{P}_{m\rightarrow n} =  \mathcal{M}'_n \approx \mathcal{M}_n,
\label{eq:plugin}
\end{equation}
in which $\oplus$ denotes architecture-level aggregation, and $\mathcal{M}'_n$ is the synthesized model used for inference with $r=n$.

However, due to the extensive range of the choices in teams of compression degree (\eg, \( r \in \{0, 1, ..., 16\} \) in DeiT-B for ToMe), training and storing such plugins for all possible compression degree pairs (\ie, $16\times17$ choices of $(m,n)$ in Eq.~\ref{eq:plugin}) would result in significant training and storage overheads. We address this issue from three perspectives.

First, inspired by parameter-efficient tuning, which uses lightweight modules to describe the gap between pre-trained and fine-tuned models, we also adopt parameter-efficient modules to describe the gap between models with different compression degrees. Concretely, we use LoRA~\cite{lora} as  $\mathcal{P}_{m\rightarrow n}$, \ie, for all weight matrix $\mathbf{W}\in\mathbb{R}^{d_1\times d_2}$ in the $\mathcal{M}'_n$ of Eq.~\ref{eq:plugin}, we have
\begin{equation}
\begin{cases}
\mathbf{W}_{\mathcal{M}'_n} = \mathbf{W}_{\mathcal{M}_m} + s\cdot\mathbf{A}\mathbf{B},\quad&\textit{if}\ \mathbf{W} \in\{\mathbf{W}_q,\mathbf{W}_v\},\\
\mathbf{W}_{\mathcal{M}'_n} = \mathbf{W}_{\mathcal{M}_m}, \quad&\textit{otherwise},
\end{cases}
\end{equation}
in which $\mathbf{W}_q$ and $\mathbf{W}_v$ are weight metrices of query and value transformations, $s$ is a hyper-parameter, and $\mathbf{A}\in\mathbb{R}^{d_1\times h}$ and $\mathbf{B}\in\mathbb{R}^{h\times d_2}$ are LoRA modules. LoRA modules comprise only about 0.1\% of the model parameters and do not incur extra computations in inference.

\begin{figure}[tb]
  \centering
  \includegraphics[height=4.5cm]{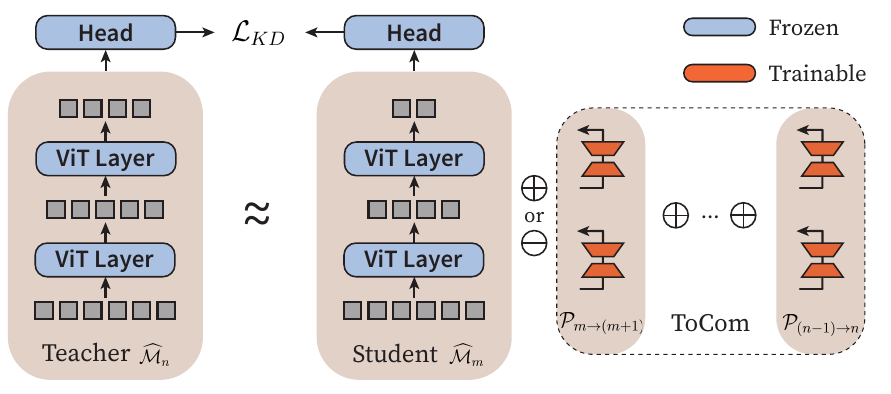}
  \caption{Illustration of our \texttt{ToCom}. \texttt{ToCom} is multiple groups of LoRA, which are trained with parameter-efficient self-distillation on pre-training dataset. The teacher model and student model have different token compression degrees which are sampled each step, and \texttt{ToCom} is plugged into student model during training.}
  \label{fig:overview}
\end{figure}

Second, we  employ LoRA to estimate the gap between models only at adjacent compression degrees, \ie, $n=m+1$ in Eq.~\ref{eq:plugin}. For the cases when $n>m+1$, we simply accumulate all the plugins between $m$ and $n$, \ie,
\begin{equation}
\mathcal{M}_m \oplus \left( \bigoplus_{i=m}^{n-1}\mathcal{P}_{i\rightarrow i+1}\right)  =  \mathcal{M}'_n \approx \mathcal{M}_n.
\label{eq:plus}
\end{equation}

Third, we assume that the gap between models is invertible, \ie, $\mathcal{P}_{n\rightarrow m}=\ominus \mathcal{P}_{m\rightarrow n}$. When $n<m$, the plugins are ``subtracted'', \ie, 
\begin{equation}
\mathcal{M}_m \ominus \left( \bigoplus_{i=n}^{m-1}\mathcal{P}_{i\rightarrow i+1}\right)  =  \mathcal{M}'_n \approx \mathcal{M}_n,
\label{eq:minus}
\end{equation}
in which $\ominus$ are the inverse of $\oplus$. To implement $\ominus$, we subtract the LoRA products $\mathbf{A}\mathbf{B}$ from the weights instead of adding it.

Now, we only need to train and store the 16 groups of LoRA plugin for ToMe to support all compression degree pairs, whose size is still negligible compared to the ViT backbone. We call the collection of these plugins \textbf{To}ken \textbf{Com}pensator (\texttt{ToCom}).  Due to the lightweight nature of \texttt{ToCom}, it can be loaded into the RAM with minimal overhead, enabling real-time inference throughput switching.


\subsection{Training \texttt{ToCom}}
As mentioned above, \texttt{ToCom} is supposed to be a universal plugin for models tuned on any downstream datasets. To enhance the generalization capability of \texttt{ToCom}, we integrate the training of \texttt{ToCom} as an extension of the pre-training stage of the ViT backbone. Specifically, we utilize the pre-training data (\eg, ImageNet~\cite{imagenet}) to train \texttt{ToCom}. 

To obtain \texttt{ToCom} supporting any compression degree pairs, we propose a self-distillation method for training it. Taking ToMe as an instance, we use $\widehat{\mathcal{M}}_n$ to denote the model obtained by \emph{directly applying ToMe with $r=n$ on the off-the-shelf pre-trained model}. The training loss is constructed as follows,
\begin{equation}
    \mathcal{L} = \begin{cases}
    \mathcal{L}_{KD}\left(\widehat{\mathcal{M}}_m \oplus \left( \bigoplus_{i=m}^{n-1}\mathcal{P}_{i\rightarrow (i+1)}\right), \widehat{\mathcal{M}}_n\right),&\textit{if}\ n > m \\
    \mathcal{L}_{KD}\left(\widehat{\mathcal{M}}_m \ominus \left( \bigoplus_{i=n}^{m-1}\mathcal{P}_{i\rightarrow (i+1)}\right), \widehat{\mathcal{M}}_n\right),&\textit{if}\ n < m
    \end{cases}
\end{equation}
in which $m$ and $n$ are randomly sampled in each training step satisfying $m\neq n$, and $\mathcal{L}_{KD}$ is the knowledge distillation loss with soft targets, \ie,
\begin{equation}
 \mathcal{L}_{KD}(\mathcal{M}_{s}, \mathcal{M}_{t})=\textrm{KL}(\mathcal{M}_{s}(\boldsymbol{x}), \mathcal{M}_{t}(\boldsymbol{x})),
\end{equation}
where $\textrm{KL}$ denote Kullback–Leibler divergence and $\boldsymbol{x}$ is the inputs. 

As illustrated in Fig.~\ref{fig:overview}, during distillation, we freeze all pre-trained parameters including the classification heads, and only update the parameter-efficient \texttt{ToCom}. It is noteworthy that while \texttt{ToCom} requires training on a large-scale dataset, it is lightweight and converges quickly. Thus, its training overhead is negligible compared to the backbone's pre-training. 

After training \texttt{ToCom}, we can directly deploy it on off-the-shelf models fine-tuned on any downstream tasks with any source degrees to any target degrees following Eq.~\ref{eq:plus} and~\ref{eq:minus}. We add or subtract \texttt{ToCom}'s LoRA products to the weight of the off-the-shelf models, which are used for inference without further training.

\begin{figure}[t]
  \centering
  \includegraphics[height=4.5cm]{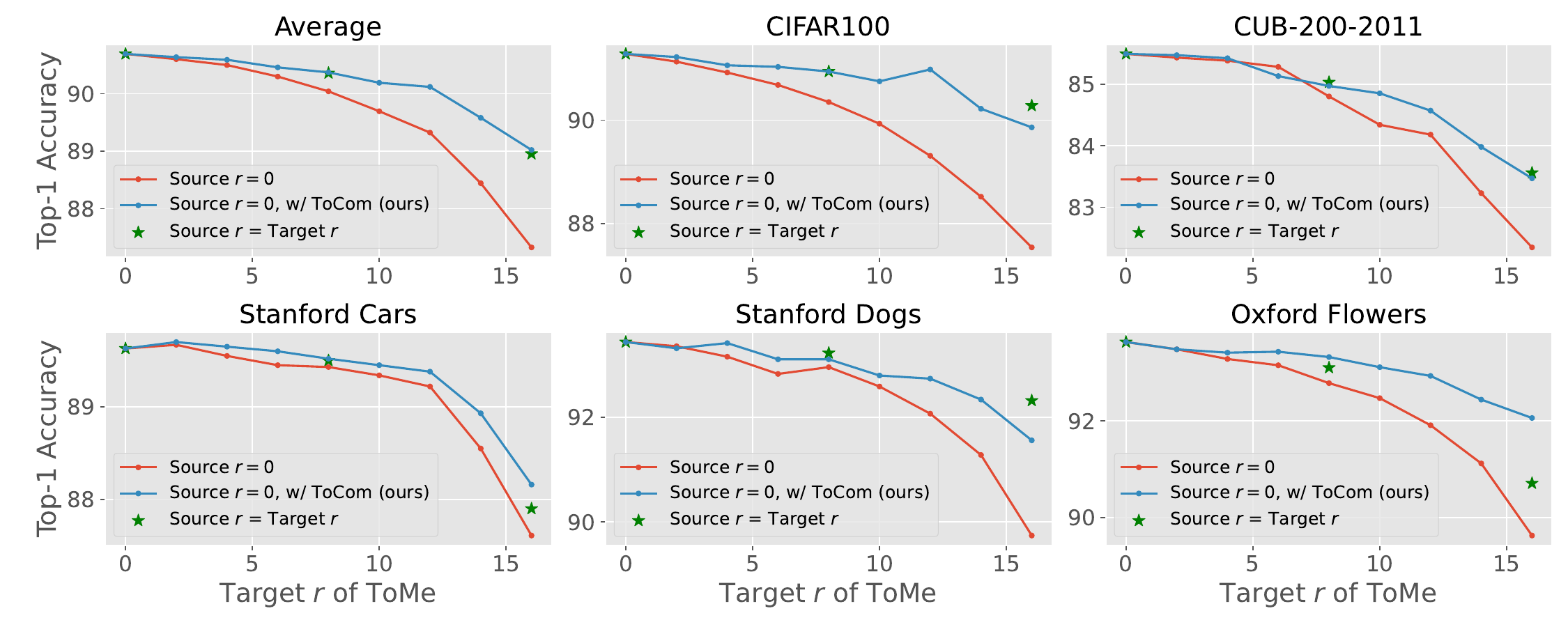}
  \caption{{Results on CIFAR100 and FGVC datasets with source $r=0$. }We also report results when source $r=$ target $r$ as reference. }
  \label{fig:fgvc}
\end{figure}

\begin{table}[t]
\begin{center}
\caption{{Results on the Natural and Specialized groups of VTAB-1k benchmark}. }
\scalebox{0.7}{
\begin{tabular}{p{2cm}<{}|p{0.75cm}<{\centering}p{0.75cm}<{\centering}p{0.75cm}<{\centering}p{0.75cm}<{\centering}p{0.75cm}<{\centering}p{0.75cm}<{\centering}p{0.75cm}<{\centering}|p{0.75cm}<{\centering}<{\centering}p{0.75cm}|p{0.75cm}<{\centering}p{0.75cm}<{\centering}p{0.75cm}<{\centering}p{0.75cm}<{\centering}|p{0.75cm}<{\centering}<{\centering}p{0.75cm}}
\toprule
&\multicolumn{9}{c|}{\textbf{Natural}}&\multicolumn{6}{c}{\textbf{Specialized}}\\
&\multicolumn{1}{c}{\STAB{\rotatebox[origin=c]{90}{Cifar100}}}
&\multicolumn{1}{c}{\STAB{\rotatebox[origin=c]{90}{Caltech101}}}
&\multicolumn{1}{c}{\STAB{\rotatebox[origin=c]{90}{DTD}}}
&\multicolumn{1}{c}{\STAB{\rotatebox[origin=c]{90}{Flower102}}}
&\multicolumn{1}{c}{\STAB{\rotatebox[origin=c]{90}{Pets}}}
&\multicolumn{1}{c}{\STAB{\rotatebox[origin=c]{90}{SVHN}}}
&\multicolumn{1}{c|}{\STAB{\rotatebox[origin=c]{90}{Sun397}}}
&\multicolumn{1}{c}{\STAB{\rotatebox[origin=c]{90}{Avg. Acc.}}}
&\multicolumn{1}{c|}{\STAB{\rotatebox[origin=c]{0}{$\Delta$}}}
&\multicolumn{1}{c}{\STAB{\rotatebox[origin=c]{90}{Camelyon}}}
&\multicolumn{1}{c}{\STAB{\rotatebox[origin=c]{90}{EuroSAT}}}
&\multicolumn{1}{c}{\STAB{\rotatebox[origin=c]{90}{Resisc45}}}
&\multicolumn{1}{c|}{\STAB{\rotatebox[origin=c]{90}{Retinopathy}}}
&\multicolumn{1}{c}{\STAB{\rotatebox[origin=c]{90}{Avg. Acc.}}}
&\multicolumn{1}{c}{\STAB{\rotatebox[origin=c]{0}{$\Delta$}}}
\\ \specialrule{0em}{1pt}{1pt}\hline\specialrule{0em}{1pt}{1pt}
\multicolumn{16}{l}{Target $r=16$ (1.9$\times$ inference speedup)}
\\ \hline
\specialrule{0em}{1pt}{1pt}
Source $r=0$&56.5&89.6&65.4&85.2&90.2&88.4&40.9&73.7&&77.1&90.1&78.6&74.4&80.0&\\
\qquad+\texttt{ToCom}&60.5&90.0&67.2&90.1&91.2&88.5&43.3&75.8&+2.1&83.7&93.6&83.0&74.3&83.6&+3.6\\
Source $r=4$&57.5&89.8&64.6&86.6&90.7&88.1&41.1&74.0&&78.9&91.6&79.9&73.8&81.0&\\
\qquad+\texttt{ToCom}&60.0&90.2&67.6&89.4&91.2&88.4&42.9&75.7&+1.7&84.5&93.7&83.2&73.7&83.8&+2.8\\
Source $r=16$&\text{\color{gray}59.6}&\text{\color{gray}89.9}&\text{\color{gray}67.5}&\text{\color{gray}89.4}&\text{\color{gray}90.7}&\text{\color{gray}89.4}&\text{\color{gray}42.0}&\text{\color{gray}75.5}&\text{\color{gray}}&\text{\color{gray}83.2}&\text{\color{gray}94.3}&\text{\color{gray}83.1}&\text{\color{gray}73.8}&\text{\color{gray}83.6}&\\ \hline\specialrule{0em}{1pt}{1pt}
\multicolumn{16}{l}{Target $r=12$ (1.5$\times$ inference speedup)}
\\ \hline\specialrule{0em}{1pt}{1pt}
Source $r=0$&60.1&90.1&68.0&89.5&91.6&89.2&42.8&75.9&&81.2&92.6&82.5&74.8&82.8&\\
\qquad+\texttt{ToCom}&61.4&90.2&68.4&91.3&91.9&89.3&43.8&76.6&+0.7&84.0&94.7&84.7&74.8&84.5&+1.7\\
Source $r=4$&60.6&90.3&67.4&90.1&91.8&88.8&42.9&76.0&&83.1&93.8&83.1&74.1&83.5&\\
\qquad+\texttt{ToCom}&61.5&90.3&68.0&91.0&92.1&88.9&43.7&76.5&+0.5&85.5&94.3&84.1&74.1&84.5&+1.0\\
Source $r=12$&\text{\color{gray}61.0}&\text{\color{gray}90.9}&\text{\color{gray}67.8}&\text{\color{gray}90.8}&\text{\color{gray}92.0}&\text{\color{gray}90.0}&\text{\color{gray}43.1}&\text{\color{gray}76.5}&\text{\color{gray}}&\text{\color{gray}84.0}&\text{\color{gray}94.6}&\text{\color{gray}84.1}&\text{\color{gray}73.7}&\text{\color{gray}84.1}&
\\
\specialrule{0em}{1pt}{1pt}
\bottomrule
\end{tabular}}\label{tab:vtab1}
\end{center}

\end{table}
\begin{table}[t]
\begin{center}
\caption{{Results on the Structured group of VTAB-1k benchmark}. }
\scalebox{0.7}{
\begin{tabular}{p{2cm}<{}|p{0.75cm}<{\centering}p{0.75cm}<{\centering}p{0.75cm}<{\centering}p{0.75cm}<{\centering}p{0.75cm}<{\centering}p{0.75cm}<{\centering}p{0.75cm}<{\centering}p{0.75cm}<{\centering}|p{0.75cm}<{\centering}p{0.75cm}<{\centering}}
\toprule
&\multicolumn{10}{c}{\textbf{Structured}}\\
&\multicolumn{1}{c}{\STAB{\rotatebox[origin=c]{90}{Clevr-Count}}}
&\multicolumn{1}{c}{\STAB{\rotatebox[origin=c]{90}{Clevr-Dist}}}
&\multicolumn{1}{c}{\STAB{\rotatebox[origin=c]{90}{DMLab}}}
&\multicolumn{1}{c}{\STAB{\rotatebox[origin=c]{90}{KITTI-Dist}}}
&\multicolumn{1}{c}{\STAB{\rotatebox[origin=c]{90}{dSpr-Loc}}}
&\multicolumn{1}{c}{\STAB{\rotatebox[origin=c]{90}{dSpr-Ori}}}
&\multicolumn{1}{c}{\STAB{\rotatebox[origin=c]{90}{sNORB-Azim}}}
&\multicolumn{1}{c|}{\STAB{\rotatebox[origin=c]{90}{sNORB-Ele}}}
&\multicolumn{1}{c}{\STAB{\rotatebox[origin=c]{90}{Avg. Acc.}}}
&\multicolumn{1}{c}{\STAB{\rotatebox[origin=c]{0}{$\Delta$}}}
\\ \specialrule{0em}{1pt}{1pt}\hline\specialrule{0em}{1pt}{1pt}
\multicolumn{11}{l}{Target $r=16$ (1.9$\times$ inference speedup)}
\\ \hline
\specialrule{0em}{1pt}{1pt}
Source $r=0$&67.6&56.2&49.4&79.6&78.6&53.3&31.8&36.8&56.7&\\
\qquad+\texttt{ToCom}&73.4&57.3&49.6&81.4&78.6&53.5&31.8&37.8&57.9&+1.2\\
Source $r=4$&70.5&55.0&49.6&81.3&77.8&53.2&33.2&36.7&57.2&\\
\qquad+\texttt{ToCom}&73.6&56.7&50.3&82.0&77.7&53.3&33.3&37.3&58.0&+0.8\\
Source $r=16$&\text{\color{gray}77.7}&\text{\color{gray}59.6}&\text{\color{gray}50.3}&\text{\color{gray}80.7}&\text{\color{gray}77.8}&\text{\color{gray}53.5}&\text{\color{gray}33.1}&\text{\color{gray}37.5}&\text{\color{gray}58.8}&\\ \hline\specialrule{0em}{1pt}{1pt}
\multicolumn{10}{l}{Target $r=12$ (1.5$\times$ inference speedup)}
\\ \hline\specialrule{0em}{1pt}{1pt}
Source $r=0$&77.1&57.6&50.2&80.5&79.0&53.7&32.3&38.5&58.6&\\
\qquad+\texttt{ToCom}&78.0&58.2&50.5&81.9&79.0&53.8&32.4&38.9&59.1&+0.5\\
Source $r=4$&78.2&56.8&50.5&82.0&77.8&53.4&33.9&37.4&58.8&\\
\qquad+\texttt{ToCom}&78.6&58.0&50.9&82.8&77.8&53.4&34.0&37.9&59.2&+0.4\\
Source $r=12$&\text{\color{gray}79.2}&\text{\color{gray}59.3}&\text{\color{gray}50.7}&\text{\color{gray}82.3}&\text{\color{gray}78.0}&\text{\color{gray}53.4}&\text{\color{gray}33.6}&\text{\color{gray}37.9}&\text{\color{gray}59.3}&\\
\specialrule{0em}{1pt}{1pt}
\bottomrule
\end{tabular}}\label{tab:vtab2}
\end{center}
\end{table}

\section{Experiments}
\subsection{Datasets and Setup}
We use ImageNet~\cite{imagenet} to train \texttt{ToCom} on pre-trained DeiT-B for 10 epochs. In \texttt{ToCom} training, we fix hyper-parameter $s=0.1$, and choose $s$ from \{0.01, 0.02, 0.05, 0.08, 0.1, 0.12, 0.15\} in inference. We use $h=2$ for each LoRA of \texttt{ToCom} such that the total number of parameter of \texttt{ToCom} remains merely 1.2M for the DeiT-B with 86M parameters. All other hyper-parameters follow the training recipe of DeiT-B. The training of \texttt{ToCom} takes only 1.8 hours on 8$\times$V100 GPUs, which is negligible compared to the 53 hours pre-training of the backbone.

We evaluate \texttt{ToCom} on more than 20 downstream datasets,  including  CIFAR100~\cite{cifar}, 4 fine-grained visual classification (FGVC) datasets, and 19 tasks in VTAB-1k benchmark~\cite{vtab}. The FGVC tasks include CUB-200-2011~\cite{cub}, Stanford Cars~\cite{cars}, Stanford Dogs~\cite{dogs}, and Oxford Flowers~\cite{flowers}. The VTAB-1k benchmark contains tasks in a large range of domains, each of which has 1000 training samples. The 19 tasks of VTAB-1k are further categorized into three groups: Natural, Specialized, and Structured. To obtain the off-the-shelf models on these downstream tasks, we fine-tune CIFAR100, FGVC, and VTAB-1k benchmarks on DeiT-B for 100, 30, and 100 epochs with batch size of 1024, 64, and 64, respectively. Moreover, since recent work has found that PET performs much better than full fine-tuning on VTAB-1k benchmark, we also apply PET when fine-tuning on VTAB-1k, \ie, freezing the pre-trained backbone and updating AdaptFormer~\cite{adaptformer} with hidden dimension 32. As for CIFAR100 and FGVC datasets, we apply full fine-tuning as default, \ie, updating all backbone parameters. Note that all these datasets are evaluated on the same \texttt{ToCom}, and no additional training is required after inserting \texttt{ToCom} into the off-the-shelf models, which means we apply pre-trained \texttt{ToCom} in a tuning-free manner across all these downstream tasks. 

\subsection{Inference-Time Acceleration on off-the-Shelf Models}
First, we evaluate \texttt{ToCom} in the scenario of inference-time acceleration, \ie, applying the models trained with lower source $r$ (or not compression) with higher target $r$ to improve the inference speed. 

On CIFAR100 and FGVC datasets, we fix source $r=0$, \ie, plug \texttt{ToCom} into models trained without ToMe, and evaluate them with  target $r\in$ \{2, 4, 6, 8, 10, 12, 16\} with inference speedup from 1.03$\times$ to 1.91$\times$. As illustrated in Fig.~\ref{fig:fgvc}, \texttt{ToCom} improves the average performance of ToMe on all target $r$. When target $r=8$ and $r=16$, \texttt{ToCom} brings 0.5 and 1.7 accuracy gains on average, respectively. Notably, when applying \texttt{ToCom},  the average off-the-shelf  performance of the model even matches that of training with source $r=$ target $r$, demonstrating the significant flexibility brought by \texttt{ToCom} to token compression.

In Table~\ref{tab:vtab1} and~\ref{tab:vtab2}, we list the results on VTAB-1k benchmark when source $r=$ 0 or 4 and target $r=$ 12 or 16.  Under the cases of source \( r = 0 \) and target \( r = 16 \), \texttt{ToCom} brings gains of 2.1\%, 3.6\%, and 1.2\% on the average accuracy of the three groups, respectively. We also provide results when source \( r \) and target \( r \) are equal as a reference. We find that in all four settings of source and target \( r \) pairs, \texttt{ToCom} consistently brings significant performance improvements, achieving performance comparable to or better than the results when source \( r \) and target \( r \). This once again validates the powerful capability of \texttt{ToCom} to bridge the gap across different \( r \) and enhance performance when accelerating inference of off-the-shelf models.

\begin{figure}[t]
  \begin{minipage}[h]{0.39\textwidth}
\centering
  \includegraphics[height=3cm]{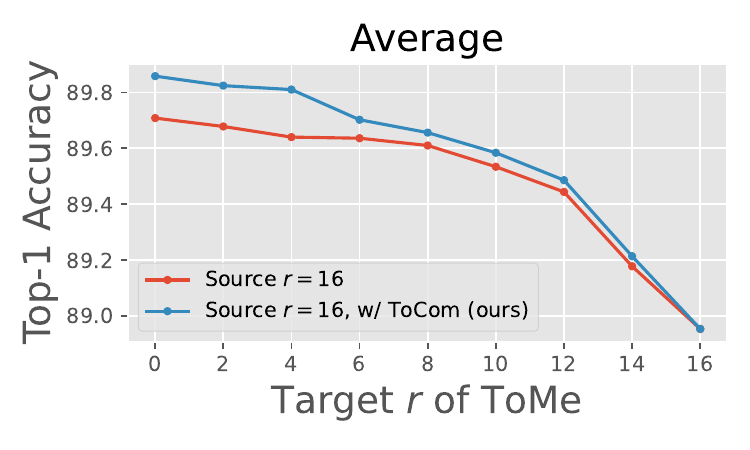}
  \captionof{figure}{{Average Results on CIFAR100 and FGVC datasets with source $r=16$. }}
    \label{fig:fgvc2}
  \end{minipage}\hfill\begin{minipage}[h]{0.59\textwidth}
\centering
\captionof{table}{{Group-wise average results on VTAB-1k benchmark. } }
\scalebox{0.7}{
\begin{tabular}{p{2cm}<{}|p{0.9cm}<{\centering}p{0.9cm}<{\centering}|p{0.9cm}<{\centering}p{0.9cm}<{\centering}|p{0.9cm}<{\centering}p{0.9cm}<{\centering}}
\toprule
&\multicolumn{2}{c}{\textbf{Natural}}&\multicolumn{2}{|c}{\textbf{Specialized}}&\multicolumn{2}{|c}{\textbf{Structured}}\\
&\multicolumn{1}{c}{\STAB{\rotatebox[origin=c]{0}{\makecell[c]{Avg.\\Acc.}}}}
&\multicolumn{1}{c}{\STAB{\rotatebox[origin=c]{0}{$\Delta$}}}
&\multicolumn{1}{|c}{\STAB{\rotatebox[origin=c]{0}{\makecell[c]{Avg.\\Acc.}}}}
&\multicolumn{1}{c}{\STAB{\rotatebox[origin=c]{0}{$\Delta$}}}
&\multicolumn{1}{|c}{\STAB{\rotatebox[origin=c]{0}{\makecell[c]{Avg.\\Acc.}}}}
&\multicolumn{1}{c}{\STAB{\rotatebox[origin=c]{0}{$\Delta$}}}
\\ \specialrule{0em}{1pt}{1pt}\hline\specialrule{0em}{1pt}{1pt}
\multicolumn{7}{l}{Source $r=16$ (1.9$\times$ training speedup)}
\\ \hline
\specialrule{0em}{1pt}{1pt}
Target $r=0$&76.0&&84.0&&57.8&\\
\qquad+\texttt{ToCom}&76.4&+0.4&84.6&+0.6&59.2&+1.4\\
Target $r=4$&76.0&&84.2&&57.9&\\
\qquad+\texttt{ToCom}&76.3&+0.3&84.6&+0.4&59.2&+1.3\\
Target $r=16$&75.5&&83.6&&58.8&\\ \hline\specialrule{0em}{1pt}{1pt}
\multicolumn{7}{l}{Source $r=12$ (1.5$\times$ training speedup)}
\\ \hline\specialrule{0em}{1pt}{1pt}
Target $r=0$&76.9&&84.5&&59.1&\\
\qquad+\texttt{ToCom}&77.0&+0.1&84.9&+0.4&59.4&+0.3\\
Target $r=4$&76.8&&84.6&&59.2&\\
\qquad+\texttt{ToCom}&76.9&+0.1&84.9&+0.3&59.5&+0.3\\
Target $r=12$&76.5&&84.1&&59.3&\\
\specialrule{0em}{1pt}{1pt}
\bottomrule
\end{tabular}    \label{tab:vtab3}
}
\end{minipage}
\end{figure}

\subsection{Training-Time Acceleration}
As illustrated in Fig.~\ref{fig:speed}, token compression also accelerates training. We here evaluate \texttt{ToCom} in the scenario when  we aim to accelerate model training without prioritizing inference speed. Specifically, we train the model with a higher source $r$ for fast training but employ a lower target $r$ (or not compression) in inference for better performance.

On CIFAR100 and FGVC datasets, we fix source $r=16$, \ie, 1.9$\times$ training speedup, and evaluate them with target $r\in$ \{0, 2, 4, 6, 8, 10, 12\}. While on VTAB-1k benchmark, we adopt source $r=$ 12 or 16 (1.5$\times$ or 1.9$\times$ training speedup) and target $r=$ 0 or 4. As illustrated in Fig.~\ref{fig:fgvc2} and Table~\ref{tab:vtab3}, the performance of models still benefits from \texttt{ToCom}. However, we notice that the improvement from \texttt{ToCom} when target $r<$ source $r$ is slighter than that when target $r>$ source $r$. Moreover, when target $r<$ source $r$, \texttt{ToCom} still exhibits a noticeable gap compared to the models trained with lower source $r$. This is because models trained with higher source $r$ receive fewer tokens with less diverse information during training which in turn leads to worse performance, despite their faster training speed. However, it is important to note that \texttt{ToCom} is a plug-and-play module that does not introduce additional training requirements or inference latency on downstream tasks. Therefore, such performance improvement at no extra cost remains meaningful.

\begin{figure}[t]
 \begin{minipage}[h]{0.49\textwidth}
\centering
\captionof{table}{Ablation study on the framework of \texttt{ToCom}. We report 19-task average results on VTAB-1k benchmark.}
\scalebox{0.7}{
\begin{tabular}{p{2.2cm}<{}|p{0.75cm}<{\centering}p{0.75cm}<{\centering}|p{0.75cm}<{\centering}p{0.75cm}<{\centering}|p{0.75cm}<{\centering}p{0.75cm}<{\centering}}
\toprule
&\multicolumn{2}{c}{{0$\rightarrow$16}}&\multicolumn{2}{|c}{{0$\rightarrow$12}}&\multicolumn{2}{|c}{{16$\rightarrow$0}}\\
&\multicolumn{1}{c}{\STAB{\rotatebox[origin=c]{0}{\makecell[c]{Avg.\\Acc.}}}}
&\multicolumn{1}{c}{\STAB{\rotatebox[origin=c]{0}{$\Delta$}}}
&\multicolumn{1}{|c}{\STAB{\rotatebox[origin=c]{0}{\makecell[c]{Avg.\\Acc.}}}}
&\multicolumn{1}{c}{\STAB{\rotatebox[origin=c]{0}{$\Delta$}}}
&\multicolumn{1}{|c}{\STAB{\rotatebox[origin=c]{0}{\makecell[c]{Avg.\\Acc.}}}}
&\multicolumn{1}{c}{\STAB{\rotatebox[origin=c]{0}{$\Delta$}}}
\\ \specialrule{0em}{1pt}{1pt}\hline\specialrule{0em}{1pt}{1pt}
Baseline&67.9&&70.1&&70.0&\\
Shared LoRA&69.2&+1.3&70.6&+0.5&70.1&+0.1\\
CLS Loss&\bf69.9&+2.0&\bf70.9&+0.8&70.6&+0.6\\
No Inversion&69.8&+1.9&\bf70.9&+0.8&70.8&+0.8\\
\texttt{ToCom} (Ours)&\bf69.9&+2.0&\bf70.9&+0.8&\bf70.9&+0.9\\
\specialrule{0em}{1pt}{1pt}
\bottomrule
\end{tabular}    \label{tab:ab1}
}
\end{minipage}
\hfill
\begin{minipage}[h]{0.49\textwidth}
\centering
\captionof{table}{Ablation study on scaling factor $s$. We report 19-task average results on VTAB-1k benchmark.}
\scalebox{0.7}{
\begin{tabular}{p{2cm}<{}|p{0.75cm}<{\centering}p{0.75cm}<{\centering}|p{0.75cm}<{\centering}p{0.75cm}<{\centering}|p{0.75cm}<{\centering}p{0.75cm}<{\centering}}
\toprule
&\multicolumn{2}{c}{{0$\rightarrow$16}}&\multicolumn{2}{|c}{{0$\rightarrow$12}}&\multicolumn{2}{|c}{{16$\rightarrow$0}}\\
&\multicolumn{1}{c}{\STAB{\rotatebox[origin=c]{0}{\makecell[c]{Avg.\\Acc.}}}}
&\multicolumn{1}{c}{\STAB{\rotatebox[origin=c]{0}{$\Delta$}}}
&\multicolumn{1}{|c}{\STAB{\rotatebox[origin=c]{0}{\makecell[c]{Avg.\\Acc.}}}}
&\multicolumn{1}{c}{\STAB{\rotatebox[origin=c]{0}{$\Delta$}}}
&\multicolumn{1}{|c}{\STAB{\rotatebox[origin=c]{0}{\makecell[c]{Avg.\\Acc.}}}}
&\multicolumn{1}{c}{\STAB{\rotatebox[origin=c]{0}{$\Delta$}}}
\\ \specialrule{0em}{1pt}{1pt}\hline\specialrule{0em}{1pt}{1pt}
Baseline&67.9&&70.1&&70.0&\\
s = 1&\bf69.9&+2.0&\bf70.9&+0.8&70.8&+0.8\\
s = 0.1 (Ours)&\bf 69.9&+2.0&\bf70.9&+0.8&\bf70.9&+0.9\\
s = 0.01&\bf69.9&+2.0&70.8&+0.7&70.8&+0.8\\
\specialrule{0em}{1pt}{1pt}
\bottomrule
\end{tabular}    \label{tab:ab2}
}
\end{minipage}
\end{figure}

\begin{figure}[t]
\begin{minipage}[h]{1\textwidth}
\centering
\captionof{table}{Ablation study on hidden dimension $h$. We report 19-task average results on VTAB-1k benchmark.}
\scalebox{0.7}{
\begin{tabular}{p{2cm}<{}|p{1.5cm}<{\centering}p{1.5cm}<{\centering}|p{1.5cm}<{\centering}p{1.5cm}<{\centering}|p{1.5cm}<{\centering}p{1.5cm}<{\centering}|p{1.5cm}<{\centering}}
\toprule
&\multicolumn{2}{c}{{0$\rightarrow$16}}&\multicolumn{2}{|c}{{0$\rightarrow$12}}&\multicolumn{2}{|c|}{{16$\rightarrow$0}}&\multirow{2}{*}{\# Params}\\
&\multicolumn{1}{c}{\STAB{\rotatebox[origin=c]{0}{Avg. Acc.}}}
&\multicolumn{1}{c}{\STAB{\rotatebox[origin=c]{0}{$\Delta$}}}
&\multicolumn{1}{|c}{\STAB{\rotatebox[origin=c]{0}{Avg. Acc.}}}
&\multicolumn{1}{c}{\STAB{\rotatebox[origin=c]{0}{$\Delta$}}}
&\multicolumn{1}{|c}{\STAB{\rotatebox[origin=c]{0}{Avg. Acc.}}}
&\multicolumn{1}{c|}{\STAB{\rotatebox[origin=c]{0}{$\Delta$}}}&
\\ \specialrule{0em}{1pt}{1pt}\hline\specialrule{0em}{1pt}{1pt}
Baseline&67.9&&70.1&&70.0&\\
h = 1&69.7&+1.8&\bf70.9&+0.8&70.8&+0.8&0.59M\\
h = 2 (Ours)& 69.9&+2.0&\bf70.9&+0.8&\bf70.9&+0.9&1.18M\\
h = 4& 69.9&+2.0&\bf70.9&+0.8&\bf70.9&+0.9&2.36M\\
h = 8&\bf 70.1&+2.2&\bf70.9&+0.8&\bf70.9&+0.9&4.72M\\
\specialrule{0em}{1pt}{1pt}
\bottomrule
\end{tabular} \label{tab:ab3}
}
\end{minipage}
\end{figure}

\subsection{Ablation Study}

To demonstrate the importance of the components of \texttt{ToCom}, we conduct ablation experiments. Firstly, to investigate the effects of parameter-efficient module arithmetic and self-distillation loss, we designed the following variants for comparison:
\begin{itemize}
    \item Shared LoRA: Unlike \texttt{ToCom}, which utilizes 16 groups of  LoRA with \( h=2 \) for arithmetic across different source-target degree pairs, we employ a single group of  LoRA with \( h=32 \), sharing it among different source-target degree pairs.
    \item CLS Loss: We use cross-entropy as classification loss instead of self-distillation loss and randomly sample $r$.
    \item No Inversion: Unlike \texttt{ToCom} which assumes $\mathcal{P}_{n\rightarrow m}=\ominus \mathcal{P}_{m\rightarrow n}$, we use two groups of LoRA for $\mathcal{P}_{n\rightarrow n+1}$ and $\mathcal{P}_{n+1\rightarrow n}$, respectively.
\end{itemize}

We conduct experiments on VTAB-1k benchmark under three setting: (source $r$, target $r$) $\in\{(0, 16), (0, 12), (16, 0)\}$. An illustrated in Table~\ref{tab:ab1}, we find that: \emph{i}) Shared LoRA performs worse than \texttt{ToCom} across  all the three settings, indicating that the LoRA parameters shared among different source-target degree pairs introduce mutual interference. \texttt{ToCom}, on the other hand, avoids this interference by using LoRA to model the gap between adjacent $r$. \emph{ii}) Classification loss performs comparably to the KD loss when \( \text{source } r < \text{target } r \). However, it exhibits poorer performance when \( \text{source } r > \text{target } r \). This is because the classification loss utilizes labels as supervision signals which is equivalent to a strong teacher, and it cannot adapt well to situations where the teacher is weak, such as when \( \text{source } r > \text{target } r \). \emph{iii}) When we allocate parameters independently for  $\mathcal{P}_{n\rightarrow n+1}$ and $\mathcal{P}_{n+1\rightarrow n}$, the performance slightly decreases despite having twice the number of parameters as \texttt{ToCom}. This indicates that there is positive transfer between  $\mathcal{P}_{n\rightarrow n+1}$ and $\ominus \mathcal{P}_{n+1\rightarrow n}$, suggesting that the assumption of $\mathcal{P}_{n\rightarrow m}=\ominus \mathcal{P}_{m\rightarrow n}$ is reasonable.

Furthermore, we investigate the impact of the key hyper-parameters of \texttt{ToCom}: LoRA hidden dimension $h$ and scaling factor $s$,  and provide results in Table~\ref{tab:ab2} and~\ref{tab:ab3}. In summary, \texttt{ToCom} is not sensitive to \( s \). The performance of \texttt{ToCom} approaches saturation when $h=2$, so we choose $h=2$ for better storage-performance trade-off. 
\begin{figure}[t]
  \begin{minipage}[h]{0.39\textwidth}
\centering
  \includegraphics[height=2.7cm]{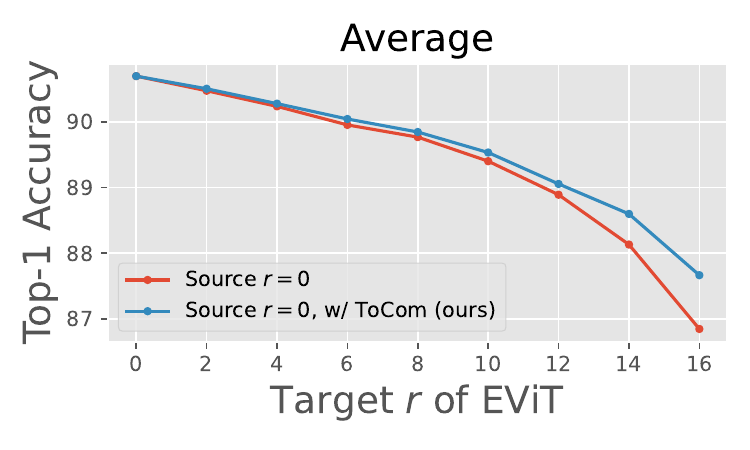}
  \captionof{figure}{{Performance of EViT on CIFAR100 and FGVC datasets with source $r=0$. }}
    \label{fig:evit}
  \end{minipage}\hfill\begin{minipage}[h]{0.59\textwidth}
\centering
\captionof{table}{{Performance of EViT  on VTAB-1k benchmark. } }
\scalebox{0.75}{
\begin{tabular}{p{2cm}<{}|p{0.9cm}<{\centering}p{0.9cm}<{\centering}|p{0.9cm}<{\centering}p{0.9cm}<{\centering}|p{0.9cm}<{\centering}p{0.9cm}<{\centering}}
\toprule
&\multicolumn{2}{c}{\textbf{Natural}}&\multicolumn{2}{|c}{\textbf{Specialized}}&\multicolumn{2}{|c}{\textbf{Structured}}\\
&\multicolumn{1}{c}{\STAB{\rotatebox[origin=c]{0}{\makecell[c]{Avg.\\Acc.}}}}
&\multicolumn{1}{c}{\STAB{\rotatebox[origin=c]{0}{$\Delta$}}}
&\multicolumn{1}{|c}{\STAB{\rotatebox[origin=c]{0}{\makecell[c]{Avg.\\Acc.}}}}
&\multicolumn{1}{c}{\STAB{\rotatebox[origin=c]{0}{$\Delta$}}}
&\multicolumn{1}{|c}{\STAB{\rotatebox[origin=c]{0}{\makecell[c]{Avg.\\Acc.}}}}
&\multicolumn{1}{c}{\STAB{\rotatebox[origin=c]{0}{$\Delta$}}}
\\ \specialrule{0em}{1pt}{1pt}\hline\specialrule{0em}{1pt}{1pt}
\multicolumn{7}{l}{Target $r=16$}
\\ \hline
\specialrule{0em}{1pt}{1pt}
Source $r=0$&73.5&&80.4&&40.8&\\
\qquad+\texttt{ToCom}&74.6&+1.1&81.7&+1.3&41.8&+1.0\\
\hline\specialrule{0em}{1pt}{1pt}
\multicolumn{7}{l}{Target $r=0$}
\\ \hline\specialrule{0em}{1pt}{1pt}
Source $r=16$&75.4&&84.2&&41.6&\\
\qquad+\texttt{ToCom}&75.7&+0.3&84.5&+0.3&43.0&+1.4\\
\specialrule{0em}{1pt}{1pt}
\bottomrule
\end{tabular}    \label{tab:evit}
}
\end{minipage}
\end{figure}
\subsection{Performance on Other Token Compression Methods}

Apart from token merging in ToMe, token pruning is also an important direction for token compression. Although previous research has found that token pruning does not perform as well as token merging on off-the-shelf models~\cite{diffrate}, we can still apply \texttt{ToCom} to token pruning to improve its performance.

We use EViT~\cite{evit}, a representative token pruning method, as the baseline. EViT uses the attention score of \texttt{[CLS]} tokens to measure the attentiveness of patch tokens and prunes inattentive ones. In the original implementation of EViT, a certain proportion of tokens are pruned from the 4th, 7th, and 11th layers of the ViT. However, we observed that gradually pruning a small number of tokens per layer, similar to ToMe, yields better performance on off-the-shelf models. Therefore, following ToMe, we also prune \( r \) tokens per layer for EViT. As illustrated in Fig.~\ref{fig:evit} and Table~\ref{tab:evit}, \texttt{ToCom} is also effective for EViT. However, we note that the performance of EViT is much worse than ToMe, especially on Structured group of VTAB-1k. This is because for some downstream tasks, changes in the number of tokens may affect the calculation of the original attention score, but ToMe mitigates this impact through its proportional attention.

\subsection{Performance on Other Backbone}

We apply \texttt{ToCom} to models with different scales, \ie, DeiT-S, and pre-training object, \ie, masked image modeling. For DeiT-S, we follow the training recipe used for DeiT-B. For ViT-B pre-trained with MAE~\cite{mae}, as it lacks a classification head, we employ L1-norm as the knowledge distillation loss at the feature level of the final layer, \ie,
$\mathcal{L}_{KD}(\mathcal{M}_{s}, \mathcal{M}_{t})=\|\mathcal{M}_{s}(\boldsymbol{x}), \mathcal{M}_{t}(\boldsymbol{x})\|_1$. As the results shown in Table~\ref{tab:deits} and~\ref{tab:mae}, \texttt{ToCom} can be generalized to different model scales and pre-training objects. In particular, because the distillation process of \texttt{ToCom} does not require labels, it can be naturally applied to various unsupervised pre-trained backbones.

\begin{figure}[t]
\begin{minipage}[h]{0.49\textwidth}
\centering
\captionof{table}{{19-task average results of DeiT-S on VTAB-1k benchmark. } }
\scalebox{0.73}{
\begin{tabular}{p{2cm}<{}|p{0.9cm}<{\centering}p{0.9cm}<{\centering}|p{0.9cm}<{\centering}p{0.9cm}<{\centering}|p{0.9cm}<{\centering}p{0.9cm}<{\centering}}
\toprule
&\multicolumn{2}{c}{\textbf{Natural}}&\multicolumn{2}{|c}{\textbf{Specialized}}&\multicolumn{2}{|c}{\textbf{Structured}}\\
&\multicolumn{1}{c}{\STAB{\rotatebox[origin=c]{0}{\makecell[c]{Avg.\\Acc.}}}}
&\multicolumn{1}{c}{\STAB{\rotatebox[origin=c]{0}{$\Delta$}}}
&\multicolumn{1}{|c}{\STAB{\rotatebox[origin=c]{0}{\makecell[c]{Avg.\\Acc.}}}}
&\multicolumn{1}{c}{\STAB{\rotatebox[origin=c]{0}{$\Delta$}}}
&\multicolumn{1}{|c}{\STAB{\rotatebox[origin=c]{0}{\makecell[c]{Avg.\\Acc.}}}}
&\multicolumn{1}{c}{\STAB{\rotatebox[origin=c]{0}{$\Delta$}}}
\\ \specialrule{0em}{1pt}{1pt}\hline\specialrule{0em}{1pt}{1pt}
\multicolumn{7}{l}{Target $r=16$}
\\ \hline
\specialrule{0em}{1pt}{1pt}
Source $r=0$&72.8&&82.2&&55.5&\\
\qquad+\texttt{ToCom}&74.3&+1.5&83.3&+1.1&56.2&+0.7\\
\hline\specialrule{0em}{1pt}{1pt}
\multicolumn{7}{l}{Target $r=0$}
\\ \hline\specialrule{0em}{1pt}{1pt}
Source $r=16$&74.3&&84.1&&57.0&\\
\qquad+\texttt{ToCom}&74.5&+0.2&84.4&+0.3&57.1&+0.1\\
\specialrule{0em}{1pt}{1pt}
\bottomrule
\end{tabular}    \label{tab:deits}
}
\end{minipage}
  \hfill\begin{minipage}[h]{0.49\textwidth}
\centering
\captionof{table}{{19-task average results of MAE-pretrained ViT-B on VTAB-1k benchmark. } }
\scalebox{0.73}{
\begin{tabular}{p{2cm}<{}|p{0.9cm}<{\centering}p{0.9cm}<{\centering}|p{0.9cm}<{\centering}p{0.9cm}<{\centering}|p{0.9cm}<{\centering}p{0.9cm}<{\centering}}
\toprule
&\multicolumn{2}{c}{\textbf{Natural}}&\multicolumn{2}{|c}{\textbf{Specialized}}&\multicolumn{2}{|c}{\textbf{Structured}}\\
&\multicolumn{1}{c}{\STAB{\rotatebox[origin=c]{0}{\makecell[c]{Avg.\\Acc.}}}}
&\multicolumn{1}{c}{\STAB{\rotatebox[origin=c]{0}{$\Delta$}}}
&\multicolumn{1}{|c}{\STAB{\rotatebox[origin=c]{0}{\makecell[c]{Avg.\\Acc.}}}}
&\multicolumn{1}{c}{\STAB{\rotatebox[origin=c]{0}{$\Delta$}}}
&\multicolumn{1}{|c}{\STAB{\rotatebox[origin=c]{0}{\makecell[c]{Avg.\\Acc.}}}}
&\multicolumn{1}{c}{\STAB{\rotatebox[origin=c]{0}{$\Delta$}}}
\\ \specialrule{0em}{1pt}{1pt}\hline\specialrule{0em}{1pt}{1pt}
\multicolumn{7}{l}{Target $r=16$}
\\ \hline
\specialrule{0em}{1pt}{1pt}
Source $r=0$&60.2&&75.3&&58.5&\\
\qquad+\texttt{ToCom}&60.7&+0.5&76.6&+1.3&58.9&+0.4\\
\hline\specialrule{0em}{1pt}{1pt}
\multicolumn{7}{l}{Target $r=0$}
\\ \hline\specialrule{0em}{1pt}{1pt}
Source $r=16$&55.1&&79.3&&57.8&\\
\qquad+\texttt{ToCom}&55.3&+0.2&79.5&+0.2&58.1&+0.3\\
\specialrule{0em}{1pt}{1pt}
\bottomrule
\end{tabular}    \label{tab:mae}
}
\end{minipage}
\end{figure}

\section{Conclusion}

In this paper, we explore the drawbacks of transformer token compression methods applied to downstream tasks. Specifically, we note that the performance of the model is suboptimal when the compression degrees during training and inference are unequal. Based on preliminary experimental results, we speculate that models trained under different compression degrees exhibit a transferable gap between tasks. To address this issue, we propose \texttt{ToCom}, a lightweight pre-trained plugin designed as a plug-and-play compensator to fill this gap and enhance the performance of token compression methods in off-the-shelf models. The effectiveness of \texttt{ToCom} is demonstrated across more than 20 downstream tasks. We believe that the potential of token compression methods has been underestimated in the past, and through \texttt{ToCom}, broader application prospects can be realized.

\bibliographystyle{splncs04}
\bibliography{main,app}
\newpage

\section*{Appendix}

\section*{Performance on Dense Prediction Tasks}
We also extend \texttt{ToCom} to semantic segmentation tasks. Since semantic segmentation needs to perform predictions on all tokens, we merge tokens before the FFN layers, and unmerge them after FFNs. We conduct experiments on ADE20k~\cite{ade} datasets using pre-trained DeiT-B$_{384}$ as encoder and Mask Transformer of Segmenter~\cite{seg} as decoder. We sample $r\in\{0, 1, 2, ..., 16\}$ in training and merge $32r$ tokens before FFNs. Since the image resolution of ADE20k is 512 $\times$ 512, we resize the image of ImageNet to 512 $\times$ 512 when training \texttt{ToCom}, and reduce the number of training epoch to 2. We provide results of single-scale evaluation in Table~\ref{tab:seg} with source $r=0$ and target $r=8,12,16$. \texttt{ToCom} also improves the performance of token compression on dense prediction tasks.

\begin{table}[h]

\begin{center}
\caption{{Performance on ADE20k validation set. We keep source $r=0$. We also report GFLOPs of encoder.}\label{tab:seg}
}
\setlength{\tabcolsep}{0.3pt}
\scalebox{0.9}{

\begin{tabular}{p{3.8cm}<{\centering}p{2.7cm}<{\centering}p{2.7cm}<{\centering}}
\hline
\specialrule{0em}{1pt}{1pt}
Setting& mIoU (SS) &GFLOPs
\\\hline\specialrule{0em}{1pt}{1pt}
Target $r=0$ & 48.7 & 106.2
\\\hline\specialrule{0em}{1pt}{1pt}
Target $r=8$ & 48.0 & 91.8
\\ 
\quad+ \texttt{ToCom}&\bf48.3& 91.8
\\\hline\specialrule{0em}{1pt}{1pt}
Target $r=12$&46.4&84.5
\\
\quad+ \texttt{ToCom}&\bf47.2&84.5
\\\hline\specialrule{0em}{1pt}{1pt}
Target $r=16$&41.3&77.3
\\
\quad+ \texttt{ToCom}&\bf43.4&77.3
\\
\bottomrule
\end{tabular}
}
\end{center}

\end{table}

\section*{Pseudocode of Training}
See Algorithm~\ref{algo}.
\begin{algorithm}
\caption{Training \texttt{ToCom}}\label{alg:cap}
\begin{algorithmic}
\Require Pre-trained model $\widehat{\mathcal{M}}$, pre-training dataset  ${\mathcal{D}}$
\State Initialize $\mathcal{P}=\{\mathcal{P}_{0\rightarrow 1},...,\mathcal{P}_{15\rightarrow 16}\}$
\For {$\boldsymbol{x} \in \mathcal{D}$}
\State Ramdomly sample $m,n$, \emph{s.t.} $m\neq n$
\State $\widehat{\mathcal{M}}_m \leftarrow$  Apply ToMe with $r=m$ to  $\widehat{\mathcal{M}}$
\State $\widehat{\mathcal{M}}_n \leftarrow$  Apply ToMe with $r=n$ to  $\widehat{\mathcal{M}}$
\If{$m<n$}
    \State $\mathcal{L} = \mathcal{L}_{KD}\left(\widehat{\mathcal{M}}_m \oplus \left( \bigoplus_{i=m}^{n-1}\mathcal{P}_{i\rightarrow (i+1)}\right), \widehat{\mathcal{M}}_n\right)$
\Else
    \State $\mathcal{L} =  \mathcal{L}_{KD}\left(\widehat{\mathcal{M}}_m \ominus \left( \bigoplus_{i=n}^{m-1}\mathcal{P}_{i\rightarrow (i+1)}\right), \widehat{\mathcal{M}}_n\right)$
\EndIf
\State Update  $\mathcal{P}$ according to $\mathcal{L}$
\EndFor
\end{algorithmic}
\label{algo}
\end{algorithm}

\section*{Experimental Details}
\subsection*{PET for VTAB-1k}
On VTAB-1k benchmark, we use PET method instead of full fine-tuning to obtain off-the-shelf models, since previous work~\cite{vpt,fact,ssf,revisit} has found that PET performs much better than full fine-tuning on VTAB-1k. Especially, \cite{revisit} finds that Adaptformer~\cite{adaptformer} outperforms other PET methods, and thus we also adopt it in experiments on VTAB-1k. AdaptFormer uses bottleneck FFN with in-between ReLU activation as adapters. The weights of an adapter are $\boldsymbol{W}_{down}\in \mathbb{R}^{d\times 32}$ and $\boldsymbol{W}_{up}\in \mathbb{R}^{32\times d}$. Adapters are inserted into networks as shortcuts of the FFN blocks, \ie, given an input $\boldsymbol{X}\in \mathbb{R}^{N\times d}$, the computation of FFNs becomes
\begin{equation}\boldsymbol{X}'=\underbrace{\boldsymbol{X}+\texttt{FFN}(\boldsymbol{\texttt{LN}(X)})}_{\text{Frozen}}+\underbrace{s\cdot\texttt{ReLU}(\boldsymbol{X}\boldsymbol{W}_{down})\boldsymbol{W}_{up}}_{\text{Tuned}}\end{equation}
where $s$ is a hyper-parameter searched from \{0.01, 0.1, 1, 10, 100\} and $\texttt{LN}$ is layer normalization.

\subsection*{Datasets}
See Table~\ref{tab:dataset}. For VTAB-1k, each dataset contains 800 samples for training and 200 for validation. Following previous work~\cite{vpt,noah,ssf,fact,convpass}, after searching the hyper-parameters, we tune the pre-trained model with all the 1,000 training and validation samples and report results evaluated on test-set. For other datasets that do not have official validation set, we randomly split 10\% training example for hyper-parameters searching.
\subsection*{Pre-Trained Backbones}
See Table~\ref{tab:pt}.

\begin{table}[h]

\begin{center}
\caption{{Pre-Trained backbones.}
\label{tab:pt}
}
\setlength{\tabcolsep}{0.3pt}
\scalebox{0.9}{

\begin{tabular}{p{3.8cm}<{\centering}p{2.7cm}<{\centering}p{2cm}<{\centering}}
\hline
\specialrule{0em}{1pt}{1pt}
Model&\makecell[c]{Pre-Training\\Dataset}&\makecell[c]{Pre-Trained\\Weights}
\\\hline\specialrule{0em}{1pt}{1pt}
DeiT-B/16~\cite{deit}&ImageNet-1K&\href{https://dl.fbaipublicfiles.com/deit/deit_base_patch16_224-b5f2ef4d.pth}{checkpoint}
\\
DeiT-S/16~\cite{deit}&ImageNet-1K&\href{https://dl.fbaipublicfiles.com/deit/deit_small_patch16_224-cd65a155.pth}{checkpoint}
\\
ViT-B/16 (MAE)~\cite{mae}&ImageNet-1K&\href{https://dl.fbaipublicfiles.com/mae/pretrain/mae_pretrain_vit_base.pth}{checkpoint}
\\
DeiT-B/16$_{384}$~\cite{deit}&ImageNet-1K&\href{https://dl.fbaipublicfiles.com/deit/deit_base_distilled_patch16_384-d0272ac0.pth}{checkpoint}\\
\bottomrule
\end{tabular}
}
\end{center}

\end{table}

\subsection*{Code Implementation}
We use \href{https://pytorch.org/}{\emph{PyTorch}} and \href{https://rwightman.github.io/pytorch-image-models/}{\emph{timm}} to implement all experiments on 8$\times$V100 GPUs.

\begin{table}[t]
\centering
\caption{
 Hyper-parameters.\label{tab:hp1}
}
\scalebox{0.9}
{
\begin{tabular}{lcccc}
\toprule
Methods & ImageNet  & CIFAR & FGVC & VTAB-1k \\
\midrule
Epochs   & 10 & 100&30&100     \\
\midrule
Batch size & 1024 & 1024&64&64\\
Optimizer & AdamW & AdamW& AdamW& AdamW\\
     Base learning rate       & 1e-3&5e-5&\{5e-3,2e-3,1e-3,5e-4,2e-4,1e-4\}&-  \\
     Learning rate&-&-&-&1e-3\\
     Learning rate decay & cosine & cosine& cosine & cosine  \\
     Weight decay        & 0.05    & 0.05 & 0.05  &0.0001  \\
     Warmup epochs  & 0 & 10 & 5 & 0      \\
     Label smoothing $\varepsilon$ & - & 0.1& 0.1& -     \\
     Stoch. Depth & 0.1 & 0.1& 0.1 & 0.1 \\
     Rand Augment  & 9/0.5        & 9/0.5& 9/0.5&- \\
     Mixup prob.  & 0.8 & 0.8& 0.8& -     \\
     Cutmix prob.   & 1.0 & 1.0& 1.0 & -    \\
     Erasing prob.    & 0.25 & 0.25& 0.25 & -   \\
 \bottomrule
\end{tabular}}

\end{table}

\begin{table*}[t]

\begin{center}
\caption{{Statistics of used datasets.\label{tab:dataset}}
}
\scalebox{0.9}{
\begin{tabular}{clcccc}
\toprule[1.5pt]
                          & Dataset              & \# Classes & Train                    & Val   & Test            \\ \midrule
                          \multicolumn{6}{c}{VTAB-1K~\cite{vtab}}\\\midrule
\multirow{7}{*}{Natural} & CIFAR100~\cite{krizhevsky2009learning}             & 100       & \multirow{7}{*}{800/1,000}                 & \multirow{7}{*}{200}   & 10,000           \\
                          & Caltech101~\cite{fei2004learning}           & 102       &                 &    & 6,084            \\
                          & DTD~\cite{cimpoi14describing}                 & 47        &                 &    & 1,880            \\
                          & Oxford-Flowers102~\cite{flower}    & 102       &                 &    & 6,149            \\
                          & Oxford-Pets~\cite{pets}          & 37        &                 &    & 3,669     \\
                          & SVHN~\cite{netzer2011reading}                 & 10        &                 &    & 26,032            \\
                          & Sun397~\cite{xiao2010sun}               & 397       &                 &    & 21,750           \\
\midrule\multirow{4}{*}{Specialized}                          & Patch Camelyon~\cite{Veeling2018qh}       & 2         &      \multirow{4}{*}{800/1,000}                 & \multirow{4}{*}{200}                & 32,768           \\
                          & EuroSAT~\cite{helber2019eurosat}              & 10        &                 &    & 5,400            \\
                          & Resisc45~\cite{cheng2017remote}             & 45        &                 &    & 6,300            \\
                          & Retinopathy~\cite{kaggle2015retinopathy}         & 5         &                 &    & 42,670           \\
 \midrule\multirow{8}{*}{Structured}                         & Clevr/count~\cite{johnson2017clevr}          & 8         &      \multirow{8}{*}{800/1,000}                 & \multirow{8}{*}{200}                  & 15,000       \\
                          & Clevr/distance~\cite{johnson2017clevr}       & 6         &                 &    & 15,000     \\
                          & DMLab~\cite{beattie2016deepmind}                & 6         &                 &    & 22,735           \\
                          & KITTI-Dist~\cite{geiger2013vision}           & 4         &                 &    & 711   \\
                          & dSprites/location~\cite{matthey2017dsprites}    & 16        &                 &    & 73,728           \\
                          & dSprites/orientation~\cite{matthey2017dsprites} & 16        &                 &    & 73,728           \\
                          & SmallNORB/azimuth~\cite{lecun2004learning}    & 18        &                 &    & 12,150     \\
                          & SmallNORB/elevation~\cite{lecun2004learning}  & 18         &                 &    & 12,150     \\ \midrule
                           \multicolumn{6}{c}{FGVC}\\\midrule
 & CUB-200-2011~\cite{cub}             & 200       & 5,994 & - & 5,794           \\
                          & Stanford Cars~\cite{car}        &  196      &  8,144 & -  & 8,041       \\
                          & Oxford Flowers102~\cite{flower}    & 102       & 1,020 & 1,020  & 6,149            \\
                          & Stanford Dogs~\cite{dogs}        &  129      & 12,000 & -  & 8,580       \\\midrule
                          \multicolumn{6}{c}{Others}\\\midrule & CIFAR100 ~\cite{krizhevsky2009learning}             & 100       &     50,000             &  -  & 10,000    \\
                          &ImageNet~\cite{imagenet}             & 1,000        &     1,281,167    & 50,000   & -   \\
                          &ADE20k~\cite{ade}             & 150        &     20,210    & 2,000   & 3,352   \\
\bottomrule[1.5pt]
\end{tabular}}\end{center}
\end{table*}

\subsection*{Data Augmentation}

\subsubsection{ImageNet \& CIFAR100 \& Few-shot learning} Following \cite{deit}, for training samples, we use color-jitter, RandAugmentation, and repeated augmentation; for validation/test samples, we resize them to $256\times256$, crop them to $224\times224$  at the center, and then normalize them with ImageNet's mean and standard deviation.

\subsubsection{VTAB-1K} Following \cite{vpt}, we just resize the images to $224\times224$.

\subsubsection{Semantic Segmentation} We completely follow the setting used in~\cite{seg}, which does mean substraction, random
resizing, random left-right flipping, and randomly crops large images and pad small images to $512\times 512$.

\subsection*{Hyper-parameters}
See Table~\ref{tab:hp1}.

\end{document}